\documentclass[graybox]{svmult}

\usepackage{amsmath}
\usepackage{array, caption, floatrow, tabularx, makecell, booktabs}
\usepackage{hyperref}
\usepackage{color}
\usepackage{colortbl}
\usepackage[colorlinks]{}
\usepackage{subfig}
\usepackage{wrapfig}
\usepackage{mathptmx}       
\usepackage{helvet}         
\usepackage{courier}        
\usepackage{type1cm}        
\usepackage{graphicx}
\usepackage[table,xcdraw]{xcolor} 

\usepackage{makeidx}         
\usepackage{graphicx}        
\usepackage{multicol}        
\usepackage[bottom]{footmisc}

\newcommand{\probP}{\text{I\kern-0.15em P}}

\makeindex             

\begin{document}
\definecolor{light-gray}{gray}{0.9}

\title*{Controlling Personality Style in Dialogue with Zero-Shot Prompt-Based Learning}
\author{\bf Angela Ramirez, \bf Mamon Alsalihy, \bf Kartik Aggarwal, \bf Cecilia Li, \bf Liren Wu, and \bf Marilyn Walker} 
\authorrunning{Angela Ramirez et. al, UCSC}
\institute{Natural Language and Dialogue Systems Lab, University of California, Santa Cruz
\email{aramir62, malsalih, kartik ,yli331,lwu35,mawalker@ucsc.edu}}
\maketitle

\abstract{Prompt-based or in-context learning has been shown  to  achieve high zero-shot performance on many natural language generation (NLG) tasks. Here we explore the performance of prompt-based learning for simultaneously controlling the personality and the semantic accuracy of an NLG for task-oriented dialogue.
We experiment with prompt-based learning on the {\sc personage} restaurant recommendation corpus  to generate semantically and  stylistically-controlled text for 5 different Big-5 personality types: agreeable, disagreeable, conscientious, unconscientious, and extravert. We test two different classes of discrete prompts to generate utterances for a particular personality style: (1) prompts that demonstrate generating directly from a meaning representation that includes a personality specification; and  (2) prompts that rely on  first converting the meaning representation to a textual pseudo-reference, and then using the pseudo-reference in a textual style transfer (TST) prompt. In each case,  we show that we can vastly improve performance by over-generating outputs and ranking them,  testing several ranking functions based on  automatic metrics for  semantic accuracy,  personality-match, and fluency.
We also test the effect of providing examples of multiple personalities, and of different sampling strategies
and  numbers of examples,
as well as testing whether NLG personality demonstrations from the restaurant domain can be used with meaning representations for the video game domain to generate personality stylized utterances about video games.  
 Our findings show that the TST prompts produces the highest semantic accuracy (78.46\% for restaurants and 87.6\% for video games) and personality accuracy (100\% for restaurants and 97\% for video games). Our results on transferring personality style to video game utterances are surprisingly good.
To our knowledge, there is no previous work testing the application of prompt-based learning to simultaneously controlling both style and semantic accuracy in NLG.} 

\begin{keywords}
personality, stylistic generation, task-oriented dialogue, natural language generation, style transfer,  prompt-based learning,  evaluation
 \end{keywords}
 
\vspace{-.2in}
 
\section{Introduction}
\label{intro-sec}

Over the last few years, prompt-based or in-context learning has been shown to  achieve high performance on many natural language generation (NLG) tasks \cite{bach-etal-2022-promptsource,li-etal-2022-prompt-based,liu2021pre}. Here we explore the performance of prompt-based learning for controlling both the personality and the semantic accuracy in natural language generation  for dialogue. 
We experiment with prompt-based learning on the  {\sc personage} corpus,  a stylistic benchmark dataset  for semantically-controlled NLG in the restaurant domain, with reference utterances  that vary stylistically according to linguistic profiles of Big-5 personality types
\cite{PennebakerKing99,Furnham90,MairesseWalker11,harrison2019maximizing,oraby2018neural,orabyreed2018}. The personality styles consist of   5 different Big-5 personality types: agreeable, disagreeable, conscientious, unconscientious, and extravert. 

We compare two different types of discrete prompts: (1)  Data-to-text (D2T) prompts that directly {\bf demonstrate} generating  from a meaning representation that includes a personality specification; (2) prompts that are based on textual style transfer (TST) {\bf instructions} \cite{reif-etal-2022-recipe}, that require first converting the meaning representation to a  a textual pseudo-reference. The two types of prompts  are illustrated in Table~\ref{table:prompt formats} for the {\it agreeable} Big-5 personality style. We also vary the number of demonstrations we provide, as well as whether the examples illustrate one personality or multiple personalities, and systematically examine the effect.

\begin{wrapfigure}{r}{2.0in}
 \centering
    \includegraphics[width=2.0in]{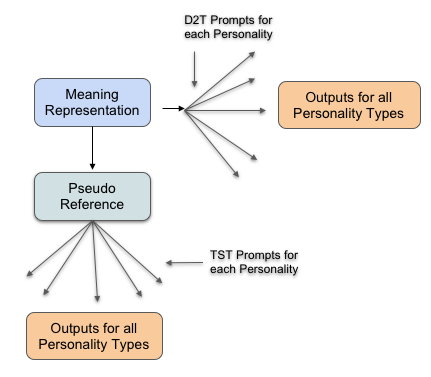}
    \caption{Two models for Semantically Controlled Generation with Style \label{model-types}}
\end{wrapfigure}
The two methods are illustrated in Figure~\ref{model-types}. Using both methods, we show that we can vastly improve performance by over-generating multiple outputs for each setting \cite{LangkildeKnight98}, and then ranking the outputs using a combination of personality accuracy, semantic accuracy and fluency. 
For semantic accuracy, we compare off-the-shelf semantic faithfulness metrics such as {\sc beyond-bleu} and {\sc bleurt} to
{\sc personage} specific scripts for calculating slot error rate \cite{wieting-etal-2019-beyond,bleurt2020,reedetal20}. To measure personality accuracy, we train a personality classifier. 

\begin{table*}[h]
\centering
\small
\begin{tabular}
{p{.95\textwidth} } 
\toprule
\multicolumn{1}{c}{ \cellcolor[gray]{0.9}  \textbf{Data To Text Prompt (D2T)} }  \\
name = nameVariable $|$ eattype = restaurant $|$ food = chinese $|$ pricerange = moderate $|$ area = riverside $|$ familyfriendly = yes $|$ near = nearVariable $|$ personality = agreeable \newline Let's see what we can find on nameVariable. oh right, it is an chinese restaurant in riverside with a quite moderate rating and it is kid friendly, also it is near nearVariable, you know, okay? 
 \newline
 \newline
 name = namevariable $|$ eattype = pub $|$ food = italian $|$ area = city centre $|$ familyfriendly = no $|$ near = nearvariable $|$ personality = agreeable
 \\
 \toprule
 \multicolumn{1}{c}{ \cellcolor[gray]{0.9}  \textbf{Textual Style Transfer Prompt (TST)} }  \\
Here is some text: \{namevariable restaurant chinese moderate riverside family friendly nearvariable\}. Here is a rewrite of the text which is agreeable : \{Let's see what we can find on nameVariable. I see it is a Chinese restaurant in riverside, also it is moderately priced and family friendly and near nearVariable.\}.
\newline
\newline
Here is some text: \{nameVariable pub Italian  city centre not family friendly nearVariable \}. Here is a rewrite of the text which is agreeable : \{ \\ 
\toprule
\end{tabular}
\vspace{-0.1cm}
\caption{Example D2T and TST prompts for the Big-5 agreeable  personality}
\label{table:prompt formats}
\end{table*}

Based on on our results for {\sc personage}, we use our best performing experimental setting  on an out-of-domain dataset for Data-to-Text NLG for Video Games, ViGGO. By doing so, we are testing whether personality examples from the restaurant domain can be used on meaning representations for the video game domain to generate personality stylized utterances about video games. The ViGGO corpus comes with a script for calculating semantic accuracy that is specific to this domain \cite{juraska2019viggo}, so we are able to apply the same ranking functions as we use for {\sc personage}.

Our results show that prompting with a single personality performs better both for achieving the target style and faithfully rendering the meaning. Our best performing setting
achieves  personality accuracies of 100\% and a best slot error rate  of 22\%.  To our knowledge, there is no previous work testing the performance of prompt-based learning for  simultaneously controlling both style and semantic accuracy in  NLG.

\section{Related Work}

Prompt-based learning  has recently been applied to  many different NLG tasks.   Previous work on semantically controlled NLG using prompt-based learning has  focused on semantic accuracy rather than attempting to simultaneously control both semantics and style \cite{reed2021,li-liang-2021-prefix, xiang2022asdot}. 
Previous work on controlling style using prompt-based learning has been framed as a textual style transfer (TST) task,  where the goal is to enhance the text with stylistic features while preserving the overall semantics and fluency of the text \cite{reif-etal-2022-recipe, suzgun2022prompt, jin-etal-2022-deep, li-etal-2022-learning-transfer}.
These measures strongly parallel the evaluation measures that we use for stylistically and semantically controlled NLG.  In TST, 
stylistic correctness is typically measured with pre-trained style classifiers, as we do here.
However it is notoriously difficult to measure semantic preservation in text-to-text tasks, where the definition of meaning tends to be quite slippery. Much work still uses  {\sc bleu} even though for many tasks it has been shown not to  correlate with human judgements
 \cite{reif-etal-2022-recipe, Papineni02bleu:a, jin-etal-2022-deep}. Newer neural measures such as {\sc beyond-bleu}, {\sc bleurt} and {\sc bertScore} have also been used, with some recent work showing that {\sc beyond-bleu} produces good results when used directly during fine-tuning \cite{liu-etal-2021-learning}.
 
 Earlier work on controlling both semantics and style was based on seq-to-seq LSTM + attention models trained with thousands of  examples \cite{oraby2018neural,oraby2019curate,tsai-etal-2021-style,reed2018can,harrison2019maximizing}. We compare our results to  previous seq-to-seq results on the {\sc personage} corpus and on the {\sc viggo} corpus 
in Section~\ref{results-sec}.

One of the key elements of our novel approach is converting our data-to-text problem to a text-to-text problem by generating pseudo-references directly from our meaning representations. Work by  Heidari et al. also experimented with different ways to convert meaning representations to textual forms for the purpose of fine-tuning \cite{heidari-etal-2021-getting}. They then show that they can use as few as 300 examples to fine-tune their NLG engine \cite{heidari2021getting}. Other work on data-to-text generation that has used  prompt-based learning has relied on models like GPT-3 to convert single KG triples into texts, and then fused those texts into a paragraph  \cite{xiang2022asdot}, but has not directly measured semantic accuracy or aimed to enforce a specific style to be generated. However this work, as well as other research, shows  that  meaning representations can be used directly in a prompt format to generate sentences \cite{soltan2022alexatm,liu2021pre,reed2021}. 
Models are clearly  sensitive to the type of prompt provided \cite{webson-pavlick-2022-prompt,bach-etal-2022-promptsource}, so we carefully compare classic data-to-text prompts with prompts that convert data-to-text to a text-to-text problem.

\section{Experimental Method}
\label{exp-method-sec}

Here, we test two different prompt-based learning approaches for semantically controlled stylistic generation, as illustrated in Figure~\ref{model-types}. 
We aim to understand which method best conditions the NLG outputs.

\subsection{Personage and ViGGO Datasets} 
\label{data-sec}

Our primary corpus is  {\sc personage}, \footnote{nlds.soe.ucsc.edu/stylistic-nlg} as illustrated in Figure~\ref{table:mr-ref-example} \cite{oraby2018neural,harrison2018neural}. {\sc personage} contains  $\sim$  88,000 restaurant recommendations  that vary along the Big Five  Personality traits: agreeable, disagreeable, conscientious, unconscientious, and extrovert \cite{Furnham90,Mairesse07,mairesse-walker-2011-controlling}.
Table~\ref{table:mr-ref-example} shows an
example MR from Personage along with five surface realizations:  a pseudo-reference generated directly from the meaning representation, as we describe in Section~\ref{SACC-sec} \cite{juraska2022}; a vanilla utterance generated for the  E2E generation challenge; examples of   the extravert, conscientious and agreeable  personality types from the {\sc personage} corpus
\cite{PennebakerKing99,Furnham90,MairesseWalker11}. 

\begin{figure}[hbt]
    \centering
    \begin{small}
    \begin{tabular}{|p{11.5cm}|}  \toprule 
          \multicolumn{1}{|c|}{ \cellcolor[gray]{0.9}  \textbf{Meaning Representation} }  \\
             \textit{name = nameVariable $|$ eatType = pub $|$ food = English $|$ priceRange = high $|$ area = city centre $|$ familyFriendly = no $|$ near = nearVariable}\\ \hline
            \multicolumn{1}{|c|}{ \cellcolor[gray]{0.9}  \textbf{Pseudo Reference} }  \\
             \textit{nameVariable pub English city centre high price range  kid friendly family friendly.}\\ \hline
          \multicolumn{1}{|c|}{ \cellcolor[gray]{0.9}  \textbf{E2E} }  \\
             \textit{nameVariable is a pub. It is an English place and it is in city centre. It has a high price range and it is kid friendly and family friendly.}\\ \hline
           
            \multicolumn{1}{|c|}{\cellcolor[gray]{0.9} \textbf{Personality: Extravert}} \\ 
          \textit{nameVariable is a pub, it is an English place and it is in city centre, it has a high price range friend and it is kid friendly and family friendly, you know!}\\
          \hline
          \multicolumn{1}{|c|}{\cellcolor[gray]{0.9} \textbf{Personality: Conscientiousness}} \\ 
          \textit{Let's see what we can find out about nameVariable. Well, it seems it isn't family friendly and it has a mediocre rating, also it's an English pub in city centre near nearVariable, also it is expensive.} \\ \hline
          \multicolumn{1}{|c|}{\cellcolor[gray]{0.9} \textbf{Personality: Agreeable}} \\ 
          \textit{You want to know more about nameVariable Oh it is an English restaurant near nearVariable and it isn't kid friendly, also it's rather expensive and a pub in city centre, okay?} \\ \hline
    \end{tabular}
\caption{Sample meaning representation with a vanilla  realization labelled E2E and three  personality-based stylistic realizations from the {\sc personage} Dataset.}
\label{table:mr-ref-example}
    \end{small}
\end{figure}

Recent work on stylistic variation in NLG for task-oriented dialogue has categorized stylistic variation into lexical, syntactic and semantic
styles \cite{tsai-etal-2021-style}. Mairesse et al. \cite{mairesse2010towards} provides a detailed summary of the psycholinguistic literature on how the Big 5 personality types are manifested in language, showing that personality affects style at all three levels, e.g. an extraverted personality will use more frequent words, will produce longer sentences with more aggregation operations, and select more positive content, while introverts tend to use rare words. The types of variation that are present in the Personage corpus are both lexical and syntactic, and categorized into Aggregation operations and Pragmatic operations. Aggregation operations modify syntactic dependency trees to combine propositions into a single sentence: these syntactic operations are aslo typically indicated by  lexical cues such  ''with'', ''and'', and
''also'', as illustrated by the Extraversion personality in Table~\ref{table:mr-ref-example}. Pragmatic operations also often involve lexical cues, but their applications typically requires  knowledge of syntactic or semantic constraints. For example,  hedges such as ``rather'' can only be placed before scalar adjectives such as ``expensive'', as illustrated by the Agreeable example in Table~\ref{table:mr-ref-example}, while insertion of hedges such as ``you know'' as shown in the Extraversion example in Table~\ref{table:mr-ref-example} is less constrained.  Adding a tag question such as ``okay?'',  or ``isn't it?" to the end of a sentencem as seen in the Agreeable example,  may require identifying the subject of the sentence in order to match the pronoun.

\begin{table}
    \small
   	\centering
   	\def\arraystretch{1.5}
    \begin{tabular}{p{11.5cm}}
        \hline
    	\rowcolor{light-gray}
    	\emph{give\_opinion}(\textsc{name} [\textbf{SpellForce 3}], \textsc{rating} [\textbf{poor}], \textsc{genres} [\textbf{real-time strategy, role-playing}], \textsc{play\-er\_per\-spec\-tive} [\textbf{bird view}]) \\
        \hline
    	I think that \textbf{SpellForce 3} is \textbf{one of the worst games} I've ever played. Trying to combine the \textbf{real-time strategy} and \textbf{role-playing} genres just doesn't work, and the \textbf{bird's eye view} makes it near impossible to play. \\
        \hline
    	\rowcolor{light-gray}
    	\emph{verify\_attribute}(\textsc{name} [\textbf{Little Big Adventure}], \textsc{rating} [\textbf{average}], \textsc{has\_multiplayer} [\textbf{no}], \textsc{platforms} [\textbf{PlayStation}]) \\
        \hline
    	I recall that you were \textbf{not that fond} of \textbf{Little Big Adventure}. Does \textbf{single-player} gaming on the \textbf{PlayStation} quickly get boring for you? \\
        \hline
    \end{tabular}
	\caption{Examples of MRs and corresponding reference utterances in the ViGGO dataset. The DA of the MRs is indicated in italics, and the slots in small caps. The slot mentions in the utterances are bolded.}
    \label{tab:ex_video_game_dataset}
\end{table}

To our knowledge,  {\sc personage} is the only corpus that provides reference utterances for  Big-5 personalities for a data-to-text generation task. However, we hypothesized
that we could achieve some style transfer to another domain by prompting with personality demonstrations from {\sc personage}, and requesting an output using a meaning
representations in another domain. Thus, after experimenting with {\sc personage}, we also test the ability to transfer
personality styles from {\sc personage} to the ViGGO video games corpus. Several examples of meaning representations and reference utterances from the original ViGGO corpus are shown in Table~\ref{tab:ex_video_game_dataset}.

\subsection{Ranking using Automatic Metrics}
\label{Automatic Metrics}

The overgenerate-and-rank method for NLG for dialogue systems assumes that overgeneration will produce multiple viable candidates, and that the best candidate(s) can be identified through ranking, in real time. In this paper, we leave aside the real-time requirement, and test whether ranking can improve the fluency, semantic accuracy, and the manifestation of personality in the selected output.  More formally,
a high-quality response generated from a model based on the personality and the  MR provided in the prompt should: (1) strongly manifest the personality; (2) have no missing or incorrect mentions of the attribute values; (3) produce no irrelevant attribute mentions i.e. hallucinations; and (4) be fluent. The generated utterance $y$, conditioned on a Personality  $P$, and an  MR $x$ with  slot values $s$, can be formulated as $y = f(P,s)$. The conditional likelihood of an utterance given P,MR can be decomposed into the product of three probabilities:

{\begin{equation}  
 \label{likelihood}
 p( y | P,s)  = p(P|y,s) * p (s| y) * p(y)
 \end{equation}
 }

The term $p(P|y,s)$ is the  probability of a particular personality given the generated utterance $y$ and the semantic attributes $s$. The  term $p(s|y)$ represents the semantic accuracy. The term $p(y)$ is the unconditional probability of the generated text. We calculate the Personality probability with a personality classifier (Sec.~\ref{personality-classifier-sec}), and test multiple ways of  computing semantic accuracy  (Sec.~\ref{SACC-sec}).
We  automatically measure the fluency of the generated text as sentence probability, as in previous work \cite{jin2022deep,suzgun2022prompt}.
We discuss how we can use these together to define several different ranking functions in Sec.~\ref{ranking-funcs-sec}..

\subsubsection{Measuring Semantic Accuracy}
\label{SACC-sec}
One advantage of 
semantically controlled natural language generation is that the meaning representation, here represented as a dialogue act with its attributes and  values, provides an  objective way to measure semantic correctness. In contrast,  work on machine translation, paraphrasing and textual style transfer (TST) use more approximate measures such as semantic similarity or {\sc bleu}, despite acknowledging {\sc bleu}'s limitations \cite{chen2011collecting,wieting-etal-2019-beyond}.

Previous work on data-to-text NLG defined a metric called the Slot Error Rate (SER) i.e. the percentage of slots $y$ that the NLG failed to realize in $x$ \cite{Wenetal15,juraska-walker-2021-attention}. 
Work on the {\sc personage} corpus defines semantic error by adding the number of substitutions S, deletions D,
repeats R, and hallucinations H \cite{harrison2019maximizing,juraska2018deep,Reedetal18,reedetal20}.\footnote{Hallucinations can only be recognized  for known attributes, but previous work has shown high correlations between human judgements and the ViGGO  and {\sc personage} SER scripts.} The SER formula is then: 
  \begin{equation}
     SER = \frac{S + D + R+  H}{N}
 \end{equation}
where N is the number of slots in the MR. Previous work on the ViGGO dataset also provides   scripts for calculating the SER  \cite{juraska-walker-2021-attention}. We use these off-the-shelf SER scripts for both
Personage and ViGGO  here.  Ranking needs an accuracy measure rather than an error measure,
so for both Personage and ViGGO, we  derive the SACC measure as:
  \begin{equation}
     SACC = 1-SER
 \end{equation}
 
\begin{wraptable}{r}{2.0in}
\small
\begin{tabular}{|l|c|c|c|}
\hline
\textbf{Measure} & \textbf{Personage} & \textbf{ViGGO}     \\ \hline
pBLEU  & 0.11 & 0.29  \\ \hline
pBeyond-BLEU & 0.29 & 0.48  \\ \hline
pBLEURT & 0.39 & 0.50   \\ \hline
pBERT precision &0.31 & 0.47  \\ \hline
pBERT recall & 0.11 & 0.34   \\ \hline
pBERT F1 & 0.24 & 0.42  \\ \hline
\end{tabular}
\caption{\label{tab:correlations} Pearson Correlation between {\sc sacc}and common Semantic Preservation Measures when applied to Pseudo References. All correlations
are statistically significant at p $=$ 0.}
\vspace{-.1in}
\end{wraptable}

However, because the {\sc sacc} metric is specific to {\sc personage} and ViGGO, we also explore the use of common
reference-based metrics for measuring semantic preservation, namely {\sc bleu}, {\sc beyond-bleu}, {\sc bleurt} and {\sc bertScore} \cite{Papinenietal02,liu-etal-2021-learning,wieting-etal-2019-beyond,bleurt2020,zhang2019bertscore}.
These metrics are designed to be reference-based, i.e. they compare a generated text with a reference text (or a set of them), typically written by humans. However we apply a novel method to  use reference-based metrics with pseudo-references, where  we compare the generated utterances directly against the input MR that has been
converted into a textual representation  \cite{juraska2022,heidari-etal-2021-getting}. The MR is linearized
and the slot values are concatenated, except that boolean-valued slots such as ``family friendly'' are represented by their slot names rather
than their values.  A {\sc personage} example  is shown in the second row of Figure~\ref{table:mr-ref-example}. While these  metrics may produce rather low scores on  (pseudo-reference, reference) text pairs, here we are interested in  relative scores rather than  absolute values. 

We assume that our domain specific {\sc sacc} measures are the best possible measure of semantic accuracy
and compute the correlations between {\sc sacc} and the reference-based metrics for both {\sc personage} and
ViGGO as shown in Table~\ref{tab:correlations}. For both {\sc personage} and ViGGO, p{\sc bleu} is the least
correlated measure while {\sc beyond-bleu}, {\sc bleurt} and {\sc bertScore} Precision are the most highly correlated.
We use these correlations in defining the ranking functions in Section~\ref{ranking-funcs-sec}.

\subsubsection{Personality Style Classifier} 
\label{personality-classifier-sec}

\begin{wraptable}{r}{2.2in}
\small
\def\arraystretch{1.5}
\begin{tabular}{|l|p{.3in}|p{.3in}|p{.3in}|}
\hline
\textbf{Personality} & \textbf{F1}  & \textbf{P}  & \textbf{R}   \\ \hline
extravert & 0.99 & 1.00 & 0.99  \\ \hline
agreeable & 0.99 & 0.99 & 0.99 \\ \hline
disagreeable & 0.99 & 0.99 & 0.99  \\ \hline
conscientious & 0.99 & 0.99 & 0.99 \\ \hline
unconscientious & 0.99 & 1.00 & 1.00  \\ \hline
\end{tabular}
\caption{\label{pers-class-acc-tab} Precision, Recall and F1 for the  Personality Classifier }
\end{wraptable}
We also need an automatic method for measuring stylistic strength or the manifestation of personality.   Style classifiers are usually  used to measure style strength in TST \cite{liu-etal-2021-learning,jin2022deep}, so we train a multi-class personality style classifier on a balanced set of  4,000 samples of reference utterances from the {\sc personage} dataset, i.e. 800 per personality \cite{reed2018can}. The classifier is a 110 million parameter BERT model that was fine-tuned with the following hyperparameters: learning rate: 3e-4, train batch size: 128, evaluation batch size: 32, epochs: 3. 
Table \ref{pers-class-acc-tab}. reports the F1, Precision, and Recall for each personality type on the {\sc personage} test set references of 1390 examples. At inference time, each sample is a stylistic personality realization generated by Jurassic.
When the personality matches the intended personality,  the  probabilities of  classifier are  used  as a term in the ranking functions  detailed in Section~\ref{ranking-funcs-sec} below.

\subsubsection{Ranking Functions}
\label{ranking-funcs-sec}

\begin{wraptable}{r}{2.2in}
\vspace{-.2in}
    \begin{small}
  	\centering
  	\def\arraystretch{1.5}
    \begin{tabular}{p{2.0in}}
    \hline
    	RF1: SACC * PAC * P(S) \\
        \hline
      \rowcolor{light-gray} 
     RF2: SACC * PAC  *P(S) * pBLEU \\
        	   \hline
     RF3: pBBLEU * PAC  *P(S) \\
        	   \hline
      \rowcolor{light-gray}
    
    	RF4: pBLEURT * PAC * P(S) \\
        \hline
    	RF5: pBERT * PAC * P(S) \\
        \hline
     \end{tabular}
     \end{small}
    \caption{Ranking Functions  }
    \label{ranking-functions-table}
\vspace{-.1in}
\end{wraptable}
Over-generate and rank is an NLG paradigm that has been around since the beginning of statistical NLG \cite{LangkildeKnight98,BangaloreRambow2000}, where multiple candidates S$_i$ are first generated, and then ranked to select the best candidate. Using a single ranking measure at a time can be misleading when
the aim is to optimize multiple aspects of the output  \cite{krishna2020reformulating}. On the other
hand, defining appropriate ranking functions is challenging, since different aspects of the output
can be weighted differently. Here we treat all aspects as equally important by multiplying our measures, and
examine the effect on the output quality.

Automatic metrics that can be combined for ranking at run time  includes measures of semantic accuracy,  the  personality classifier probabilities (PAC), and fluency, calculated as  P(S),  the probability of candidate S according to an LLM. We experiment with the five ranking functions in Table~\ref{ranking-functions-table}. RF1 captures all three components of the desired output, personality match (PAC), semantic accuracy (SAC) and fluency P(S). Because p{\sc bleu} is the least correlated metric with {\sc sacc}, we add it as an extra term to the ranking function in RF2 to see whether enforcing lexical similarity  improves results. The other three measures all explore the effect of replacing
{\sc sacc} with a general measure for semantic similarity. As shown in Table~\ref{tab:correlations},
{\sc beyond-bleu}, {\sc bleurt} and {\sc bertScore} Precision are the most highly correlated with {\sc sacc}, so we replace
{\sc sacc} with these three measures in RF3, RF4 and RF5. We calculate P(S) using GPT2-Large \cite{radford2019language}. 

\subsection{Prompt Formats, Prompt Sampling, and Prompt Selection}
\label{prompt-format-sec}

Our experiments test different prompt formats, number of examples, and sampling methods. We test two discrete prompt formats, using the traditional Data-to-Text representations (D2T) as well as representations similar to those used for Textual Style Transfer (TST). The two formats were shown in Table~\ref{table:prompt formats}, namely D2T format demonstrates generating an utterance directly from the meaning representation with a personality token, while the TST format provides instructions to generate a particular personality   from a textual pseudo-reference of the meaning representation. 

Prompt-based learning is restricted by the number of input examples provided: for Jurassic the maximum  is about 36 examples.  In addition to varying the prompt format, we also experiment with the number of examples, as previous work has shown that  also matters \cite{soltan-etal-2021-limitations}. We test between 1 and 36 examples.

We also experiment with whether we can create a ``control knob" for personality by presenting all 5  types of personalities in some prompts and only a single personality in other prompts.  Table \ref{table:prompt formats} provides an example of one personality being used for few shot learning. Experiments using only one personality at a time used either 10 examples or 36 examples. Experiments using all personalities used either 1 example per personality (5 total examples) or  6 examples per personality (30 total examples). In these experiments,  examples were randomly selected from the original {\sc personage} train set where we select examples given the criteria of the number of examples and the number of personalities. 

In addition, once we determine a good setting for type of prompt and number and type of examples, we
build on previous work using a diversity criterion for selecting prompts for instruction tuning \cite{wang2022self}. 
We hypothesized that creating prompt examples using a diversity criteria might
lead to better performance. While the instruction tuning work  used ROUGE score to select diverse automatically generated prompts, we use {\sc bleurt}, and select a set of prompts that are the least similar according to {\sc bleurt}. 
We start with a large random sample from the training set, and randomly select our first example.
We then calculate the {\sc bleurt} score between our first example and all the other examples in our random sample. We greedily
then select a second example  with the lowest {\sc bleurt} score, i.e. with the lowest similarity to the example already in the pool. We repeat this process, comparing each new candidate to the examples already in the pool, and selecting the one with the lowest average {\sc bleurt}  until we reach n number of examples, $T = {e_1, e_2 ... e_n}$.  Experiments using this selection process will be called \textit{diverse}. 

Finally,  based on our findings on {\sc personage}, we utilize the best experimental settings to attempt personality style transfer in an out-of-domain dataset for Video Games called ViGGO, by providing demonstrations of personalities in the restaurant domain, with test items from ViGGO test set. \cite{juraska2019viggo}.

\section{Results}
\label{results-sec}

All experiments use  the Jurassic-1 Jumbo 175B parameter PLM, a publicly available 178B parameter autoregressive language model \cite{lieberetal21,levine2020}. Based on  tuning experiments with different settings,  we set temperature at .7 and top P at 1.

We first report results comparing the two prompt formats and types of samples for {\sc personage},
and then we report results for the five ranking functions. Finally we report results for
personality transfer on the ViGGO corpus, and present a qualitative analysis of personality expression and diversity.

\vspace{.1in}
\noindent{\bf Prompt Style and Prompt Sampling.}
Here we compare the two prompt formats, Data-to-Text (D2T) and Textual Style Transfer (TST) for {\sc personage}, when providing examples of either a single personality or examples of all 5 personalities, and varying  the number of examples per prompt. 
Table~\ref{table:prompt formats} provided examples of both the D2T and TST prompt styles and
Table\ref{tab:d2t-tst-experiments} provides the experimental results. 

\begin{table}[h!tb]
\renewcommand*{\arraystretch}{1.2}
\begin{tabular}{|c|p{.55in}|p{.55in}| p{.55in}|p{.55in}|p{.55in}|p{.55in}|}
\hline
\bf ID &	\bf SACC BR & \bf	SACC AR &	\bf PAC BR  & \bf	PAC AR  &	\bf Perfect BR & \bf  Perfect  AR \\ \hline \hline
 \rowcolor{light-gray} \bf D2T-10-specific&	66.08\%   & 	65.77\%   & 	97.61\%   & 	100.00\%   & 	6.80\%   & 	13.45\%    \\ \hline
 \rowcolor{light-gray} \bf D2T-36-specific&	68.26\%   & 	68.50\%   & 	19.68\%   & 	88.27\%   & 	1.83\%   & 	9.86\%    \\ \hline
 \rowcolor{light-gray} \bf D2T-1-all&	63.24\%   & 	62.12\%   & 	19.53\%   & 	86.26\%   & 	0.98\%   & 	5.90\%    \\ \hline
 \rowcolor{light-gray} \bf D2T-6-all&	69.03\%   & 	68.48\%   & 	19.68\%   & 	88.06\%   & 	1.91\%   & 	11.29\%    \\ \hline \hline

\bf TST-10-specific&	\bf 72.02\%   & \bf	78.23\%   & 	\bf 99.00\%   & 	97.55\%   & 	7.19\%   & 	\bf 36.69\%    \\ \hline
\bf TST-36-specific&	68.07\%   & 	73.72\%   & 	97.00\%   & 	97.63\%   & 	5.19\%   & 29.21\%   \\ \hline
\bf TST-1-all&	66.36\%   & 	70.30\%   & 	38.00\%   & 	97.63\%   & 	4.32\%   & 	25.76\%    \\ \hline
\bf TST-6-all&	70.96\%   & 	75.81\%   & 	56.00\%   & 	97.63\%   & 	6.21\%   & 	35.47\%    \\ \hline \hline
\bf TST-10-diverse&	67.96\%   & \bf	78.46\%   & 	98.00\%   & \bf	100.00\%   & 	9.47\%   & \bf	38.99\%   \\ \hline 
 \end{tabular}
 \caption{Results comparing Data-to-Text prompts vs. Textual Style Transfer prompts using  RF2 for ranking.  Rows indicate both Prompt type and number of examples in prompt.    BR (Before Ranking) metrics reports the average score   over the entire candidate pool of 13900 outputs.  AR (After  Ranking) reports performance after selecting the best candidate according to RF2. {\sc sacc} $=$ Semantic Accuracy. PAC $=$ Personality Accuracy.  \textbf{Perfect} reports the percentage of candidates that are correct for both semantic accuracy and personality realization.}
\label{tab:d2t-tst-experiments}
\end{table}

The top part of Table~\ref{tab:d2t-tst-experiments} shows that
the D2T prompts consistently perform worse in every experimental setting, independently of whether examples are provided
of multiple personalities (all) or single personalities (specific) or whether fewer or more examples are provided. For example, comparing D2T-10-specific to TST-10-specific, we can see that after ranking, the TST-10-specific setting (10 examples of a specific personality) provides the best performance with a semantic accuracy of 78.23\% and a stylistic accuracy (PAC) of 99.00\%. This supports our hypothesis that textualizing the data-to-text representations to make them look more like the natural text that LLMs are trained on would result in better performance. We also achieve a significantly higher stylistic accuracy over the whole candidate pool (99.00\% PAC BR for TST-10-specific as compared to 96.71\% PAC BR for D2T-10-specific), by instructing the model to realize the content as a particular personality type, rather than just demonstrating. Also interestingly,
the D2T performance for {\sc sacc} is the same before (BR) and after (AR) ranking, while the improvements in {\sc sacc} after ranking are large for the TST prompt style. This shows that the overall quality of the pool of D2T candidates is lower.

The lower part of Table~\ref{tab:d2t-tst-experiments} focuses on the TST experiments. These results show that it is more challenging to get the LLM to learn from diverse prompts how to do more than one task at a time, i.e. performance is lower when we provide examples of all personalities, such  as TST-1-all (5 examples with one for each personality) and TST-6-all (30 examples with 6 for each personality). Moreover, interestingly, performance is better with only
10 examples of a specific personality, rather than 36 examples. The LLM may find long contexts such as would be provided with 36 examples more challenging than a shorter context.

Finally, we compare our random sampling method for our best setting to our diversity promoting sampling method described in Section~\ref{prompt-format-sec} \cite{wang2022self}. The bottom row of Table~\ref{tab:d2t-tst-experiments} shows that we get a slight improvement in SACC to 78.46\% by sampling more diverse prompts as well as an improvement from 97.55\% to 100\% for personality accuracy after ranking (PAC AR). While the difference in {\sc sacc} is not significant (p$=$ .28), the difference in PAC AR is significant (p$=$ 0). We therefore conclude that the diversity selection method is beneficial.

However compared to SOTA of 99.00\% SACC for fine-tuning, the overall performance for {\sc sacc} is low \cite{harrison2019maximizing}. Semantically and stylistically perfect outputs are also rare. Previous work comments that it is difficult for a model to simultaneously achieve high stylistic accuracy and high semantic accuracy. Here personality accuracy is high but there are few semantically perfect outputs to choose from.  

\vspace{.1in}
\noindent{\bf Ranking.} 
We now explore how different ranking functions affect the results. 
Table~\ref{tab:results-tab} provides the results for the best experimental setting for each ranking function from  
Section~\ref{ranking-funcs-sec}. Here, we only experiment with the diverse prompts, given the results in the last row of Table~\ref{tab:d2t-tst-experiments}, 

\begin{table}[htb]
\renewcommand*{\arraystretch}{1.5}
\begin{tabular}{|p{.5in}|p{1.5in}|p{0.8in}|p{.45in}|p{.45in}|p{.45in}|}
\hline
\textbf{ID} & \textbf{Formula} & \bf ID & {\textbf{SACC}} & \bf PAC & \bf  BLEU\\ \hline
 \textbf{RF1} & SACC * PAC * P(S) & TST-10-diverse  & 76.56\% &	100.00\%&	0.235   \\ \hline
 \textbf{RF2} & SACC * PAC  *P(S) * pBLEU & TST-10-diverse & \bf 78.46\%&	\bf 100.00\%&	0.240  \\ \hline
 \textbf{RF3} & pBBLEU * PAC  *P(S) & TST-10-diverse & 65.87\%&	100.00\%&	0.224\\ \hline
\textbf{RF4} &  pBLEURT * PAC * P(S) & TST-10-diverse & 71.61\%&	98.20\%	&0.213\\ \hline
\textbf{RF5} & pBERT * PAC * P(S) & TST-10-diverse & 63.10\%	&100.00\%&	0.219\\ \hline 
 \end{tabular}
 \caption{Results on {\sc personage} using all Ranking functions for prompt examples selected using a diversity criteria (TST-diverse), for the TST-10 best prompt setting. {\sc sacc} $=$ Semantic Accuracy. PAC $=$ Personality Accuracy. {\sc bleu} is Corpus {\sc pbleu}.}
\label{tab:results-tab}
\end{table}

Row RF2 of Table~\ref{tab:results-tab} shows 
that the addition of the {\sc pbleu} term to RF1 achieves   higher semantic accuracy (p$=$ 0) and higher {\sc bleu}, while maintaining the same stylistic accuracy (PAC) of 100\%.  We speculate that  the addition of the {\sc pbleu} term favors outputs whose lexical realizations more closely match the original MR, enabling the SER script to more easily identify semantically correct realizations. 

Comparing RF3, RF4 and RF5 in the last three rows of Table~\ref{tab:results-tab} shows that the best performing off-the-shelf semantic accuracy function is {\sc bleurt}, with RF4 performing significantly better than both {\sc beyond-bleu}  and {\sc bertScore} (p$=$ 0), although with somewhat lower personality accuracy.
 
\begin{table}[htb]
\renewcommand*{\arraystretch}{1.5}
\begin{tabular}{|p{.35in}|p{1.7in}|p{0.8in}|p{.45in}|p{.45in}|p{.45in}|}
\hline
\textbf{ID} & \textbf{Formula} & \bf ID & {\textbf{SACC}} & \bf PAC & \bf  BLEU\\ \hline \hline
 \rowcolor{light-gray} \bf RF1	& SACC* PAC * LMPROB&	TST-10-diverse&	86.02\%&	96.44\%&	0.139 \\ \hline
 \rowcolor{light-gray} \bf RF2	&SACC*PAC*LMPROB*PBLEU	&TST-10-diverse&	85.59\%&	96.39\%&	0.138\\ \hline
 \rowcolor{light-gray} \bf RF3	&BBLEU*PAC*LMPROB&	TST-10-diverse&	\bf 86.15\%& \bf	96.61\%&	0.139\\ \hline
 \rowcolor{light-gray} \bf RF4	&BLEURT*PAC*LMPROB&	TST-10-diverse&	87.58\%&	57.06\%&	0.168\\ \hline
 \rowcolor{light-gray} \bf RF5	&BERT*PAC*LMPROB	&TST-10-diverse	&85.59\%&	96.61\%	 & 0.138 \\ \hline \hline
\bf RF1	& SACC*PAC*LMPROB &	TST-10&	80.75\%&	96.44\%&	0.095 \\ \hline
\bf RF2	&SACC*PAC*LMPROB*PBLEU	&TST-10	&85.44\%&	96.00\%&	0.113\\ \hline
\bf RF3	&BBLEU*PAC*LMPROB&	TST-10&	84.40\%&	94.44\%&	0.107\\ \hline
\bf RF4	&BLEURT*PAC*LMPROB	&TST-10	&95.75\%&	55.94\%&	0.398\\ \hline
\bf RF5	&BERT*PAC*LMPROB	&TST-10	&80.00\%&	96.61\%& 0.063	\\ \hline
 \end{tabular}
 \caption{Results on ViGGO for all Ranking functions for prompt examples selected using a diversity criteria (TST-diverse) vs.  randomly selected prompts for the TST-10 best prompt setting. {\sc sacc} $=$ Semantic Accuracy. PAC $=$ Personality Accuracy. {\sc bleu}  here is Corpus pseudo {\sc bleu}.}
\label{tab:viggo-results-tab}
\end{table}	 

\vspace{.1in}
\noindent{\bf Out of Domain Results for ViGGO.} We now turn to our experiments using example prompts  from the {\sc personage} corpus with meaning representations for the ViGGO  corpus. Our goal is to see whether we can transfer personality style across domains. Table~\ref{tab:ex_video_game_dataset} provided examples of ViGGO MRs and vanilla outputs: there are no reference utterances for ViGGO outputs with personality. We apply the best prompt format and number of prompts combination (TST-10) from the {\sc personage} experiments on ViGGO,  for both the randomly sampled and diverse prompt sets  to test how diverse examples affects generalization across domains.

\begin{table*}[htb]
\begin{footnotesize}
\begin{tabular}
{@{} p{1.4cm}|p{5.0cm} |p{5.0cm} @{}} \toprule
\textbf{Personality} & \textbf{Generated Text} & \textbf{Reference Text}  \\ \midrule 
{\cellcolor[gray]{0.9} \sc Agree} & {\cellcolor[gray]{0.9}``{\bf Hey," I say. ``You want to know more about Marinello's? Right, right, right," I say}. ``It has {\bf like}, an average rating and it is an Italian place in city centre, {\bf also} it is a pub, {\bf you know, alright?}"} &  {\cellcolor[gray]{0.9} {\bf Let's see what we can find on Marinello's}. {\bf Yeah} it is an Italian restaurant, family friendly and expensive in the city centre, {\bf also} it has an average rating, and it is a pub {\bf pal}, {\bf okay?} {\bf you know}.} \\ \hline
 {\sc Disagree} & {\bf Oh God} it's an English pub with a {\bf relatively} moderate price range and it isn't kid friendly. it is near nearVariable. {\bf obviously}. & {\bf Damn} moderately priced nameVariable is  in city centre {\bf and} it isn't family friendly, {\bf also} it's an English place. It is near nearVariable. It is a pub. \\ \hline
{\cellcolor[gray]{0.9}{\sc Consc}} & {\cellcolor[gray]{0.9} {\bf Right, let's see what we can find on} nameVariable. {\bf I see}, it is a restaurant, {\bf you know} and it isn't kid friendly, {\bf also} it is a Italian place, {\bf also} it is in riverside, {\bf and} it has a so-so rating.} & {\cellcolor[gray]{0.9} {\bf Let's see}, nameVariable... {\bf Well, I see} nameVariable is moderately priced, an Italian restaurant near nearVariable {\bf and} kid friendly in riverside {\bf with} a mediocre rating, {\bf also} it is a pub.}  \\ \hline
 {\sc Unconsc} & {\bf Yeah, oh God I am not sure. Oh, i mean, oh God. Oh God}. {\bf Anyway}, nameVariable is a fast food restaurant, near nearVariable, {\bf also} it is in city centre, and it isn't family friendly.   &  {\bf Oh gosh mmhm... I don't know. I mean}, nameVariable is a restaurant, {\bf also} it is in city centre, it is expensive near nearVariable, {\bf also} it isn't kid friendly, and it's a fast food place. \\ \hline
{\cellcolor[gray]{0.9} {\sc Extra}} & {\cellcolor[gray]{0.9} {\bf There you are, now, let's see what we can find on} nameVariable. {\bf Well}, nameVariable is a fast food restaurant in city centre, {\bf also} it is a restaurant, it is moderately priced {\bf and} it is family friendly. } &  {\cellcolor[gray]{0.9} nameVariable is moderately priced near nearVariable and family friendly in city centre, it is a fast food place, {\bf you know buddy} and it is a restaurant{\bf ! } } \\ \hline
\end{tabular}
\vspace{-.1in}
\caption{Examples of Generated Text and References for Each Personality.   Cues indicative of each personality type are shown in bold. }
\label{table:personality-outputs}
\end{footnotesize}
\end{table*}

Table~\ref{tab:viggo-results-tab} shows that the results for ViGGO are surprisingly good, and that there is good personality transfer across domains, with personality accuracies of 97.00\%. Interestingly, the upper part of Table~\ref{tab:viggo-results-tab} shows that the diverse prompts yield higher overall {\sc sacc}, suggesting better generalization via diversity. Row 2 of Table~\ref{tab:viggo-results-tab} shows that RF3 with diverse prompts has the best combined performance for {\sc sacc} and personality accuracy. The RF4 rows in both the  parts of  Table~\ref{tab:viggo-results-tab} show that RF4, using {\sc bleurt}  provides the highest {\sc sacc}, and the highest {\sc pbleu} score, but unacceptably low  PACs of  57.00\% and 56.00\%. Perhaps the {\sc bleurt} metric ranks candidates lower that manifest personality.  A paired t-test shows that  RF3 performs significantly better than RF1 (p $=$ 0) for {\sc sacc}, even though  
   {\sc bleurt} was most highly correlated with {\sc sacc}. Again, {\sc sacc} is low compared to the fine-tuning SOTA of 99.2\% \cite{juraska-walker-2021-attention}.

\vspace{.1in}
\noindent{\bf Qualitative Analysis.} Table \ref{table:personality-outputs} provides generation outputs
for  each personality along with their reference  texts for the restaurant domain, while 
Table~\ref{table:viggo_sentences} provides Viggo generation outputs when conditioned on personality utterances from {\sc personage}.
In both tables we mark in bold  the  linguistics markers of personality for each personality type \cite{mairesse2010towards}. Table \ref{table:personality-outputs} illustrates how the LLM generalizes from the examples given for each personality to produce similar markers that would not have been seen in the demonstrations, for example the ``Hey'' formulation for agreeableness in the first row is completely novel, as is the ``There you are, ...'' formulation for extraversion in the last row. Table~\ref{table:viggo_sentences} illustrates the differences between the vanilla reference sentences for Viggo and the outputs that have been personality conditioned, with many of the basic linguistic markers appearing in the ViGGO outputs.

\begin{table*}[htb]
\begin{footnotesize}
\begin{tabular}
{@{} p{1.4cm}|p{4.75cm} | p{4.75cm}   @{}} \toprule
\textbf{Personality} & \textbf{Generated Text} & \textbf{Reference Text}   \\ \midrule

{\cellcolor[gray]{0.9} \sc Agree} &  {\cellcolor[gray]{0.9} I {\bf personally} think that Call of Duty: Advanced Warfare 2014 is an average first person shooter, with pretty good graphics, \textbf{alright?}} & {\cellcolor[gray]{0.9}  For a 2014 game, I think Call of Duty: Advanced Warfare wasn't so fun to play from the first person perspective. }  \\ \hline

 {\sc Disagree}  & \textbf{Oh God}, I recommend Assetto Corsa, it's {\bf basically} a first person Steam game, \textbf{damn near} perfect. & Assetto Corsa is a first person title on Steam. Are you familiar with it? \\ \hline
 
{\cellcolor[gray]{0.9}{\sc Consc}} &  {\cellcolor[gray]{0.9}  \textbf{Let's see what we can find on} Little Big Adventure. \textbf{Well}, {\bf it seems that} Little Big Adventure is a game and has a mediocre rating, {\bf also} it isn't multiplayer, {\bf also} it is for the PlayStation. }  &  {\cellcolor[gray]{0.9} I remember you saying you found Little Big Adventure to be average. Are you not usually that into single-player games on PlayStation?}\\ \hline
 {\sc Unconsc} &  \textbf{Mmhm... I know, er...} F1 2014 is a driving/racing game, also it is a simulation, also it is a sport multiplayer, also... & If you're looking for a multiplayer sport racing sim, you can't go wrong with F1 2014. Are you familiar with it?\\ \hline
 
{\cellcolor[gray]{0.9} {\sc Extra}} & {\cellcolor[gray]{0.9} My opinion is that Guitar Hero: Smash Hits is a poor first person game, \textbf{you know!}}  &  {\cellcolor[gray]{0.9} I wanted to like Guitar Hero: Smash Hits but honestly, it was just an underwhelming first person game.}\\ \hline
\end{tabular}
\vspace{-.1in}
\caption{ViGGO Personality Outputs with Reference utterances from  ViGGO}
\label{table:viggo_sentences}
\end{footnotesize}
\end{table*}

\section{Discussion and Conclusion}
\label{related-sec}

We tested two types of discrete prompts for stylistically and semantically controlled NLG, and show  treating data-to-text as a text-to-text task performs better for both semantic and stylistic accuracy.  To our knowledge these are the first results testing prompt-based learning for simultaneously controlling  both semantics and style.

We varied the number of prompt examples, and the sampling of examples, to either sample multiple personalities or single personalities, and comparing random sampling to a sampling to encourage diversity. We found that examples illustrating  multiple personalities in the same prompt produces worse performance rather than encouraging generalization. In addition, we find that selecting sample demonstrations using a diversity criterion improves both semantic and stylistic accuracy as well as stylistic transfer to the video games domain. 

We also surprisingly get lower overall semantic accuracies when prompting and testing with restaurant examples than we do when prompting with restaurant examples and testing with video game MRs. We speculate that this may be  due to the delexicalization of the restaurant name in the {\sc personage} corpus. {\sc personage} is based on synthetic MRs created for the E2E generation challenge. This means that the MRs do not describe real restaurants,  while the ViGGO MRs do correspond to real video games. Thus ViGGO realizations can benefit from the knowledge that the LLM has about video games. In future work, we hope to test whether using MRs that correspond to real restaurants improves semantic accuracy.

This work has several limitations. One limitation is the overall semantic accuracy performance for both restaurants and video games. These are lower than fine-tuned models and thus, in a real setting, fine-tuned models would still need to be used,  as well as the fact that the Jurassic model cannot be run in real time. Both of these limitations might be addressable by instruction tuning a smaller model for data-to-text and stylistic control tasks such as we report here \cite{ouyang-etal-2021-dialogue,wang2022self}. Another limitations is that we
only tested our approach on  two domains, and only on five personality styles.
 

\bibliographystyle{abbrv}

\end{document}